\documentclass{article}
\PassOptionsToPackage{sort,square,comma,numbers}{natbib}
\usepackage[final]{nips_2016}
\usepackage[utf8]{inputenc} 
\usepackage[T1]{fontenc}    
\usepackage{hyperref}       
\usepackage{url}            
\usepackage{booktabs}       
\usepackage{amsfonts}       
\usepackage{nicefrac}       
\usepackage{microtype}      
\usepackage{bm}
\usepackage{amsmath}
\usepackage{mathtools}
\usepackage{tabularx,graphicx}
\usepackage{bbm}
\usepackage{pdflscape}
\usepackage{rotating}
\usepackage[bottom]{footmisc}
\usepackage[font=small]{subcaption}
\DeclareCaptionFont{mdit}{\itshape}
\usepackage[font=small,textfont=mdit]{caption}
\usepackage{tabularx}
\usepackage{float}
\usepackage{empheq}
\usepackage{multicol}
\usepackage{xfrac}    
\usepackage{nicefrac} 
\usepackage{diagbox}
\usepackage{tikz}
\usepackage{graphicx}
\usepackage[rightnl]{algorithm2e}
\usepackage{times}
\usepackage{dsfont}
\usetikzlibrary{positioning}
\def\mathclap#1{\text{\hbox to 0pt{\hss$\mathsurround=0pt#1$\hss}}}
\newcommand{\myparagraph}[1]{\vspace{-1mm}\paragraph{#1}}
\newcommand{\mysection}[1]{\vspace{-3mm}\section{#1}\vspace{-2mm}}
\newcommand{\mysubsection}[1]{\vspace{-2mm}\subsection{#1}\vspace{-1mm}}
\newcommand{\mycaption}[1]{\vspace{-1.5mm}\caption{#1}\vspace{-4mm}}

\DeclareMathOperator*{\argmin}{arg\!\min}
\DeclareMathOperator*{\argmax}{arg\!\max}

\title{DISCO Nets: DISsimilarity COefficient Networks}

\author{
  Diane Bouchacourt\\
University of Oxford\\
$\texttt{diane@robots.ox.ac.uk}$
\\
\And
  M. Pawan Kumar\\
University of Oxford\\
$\texttt{pawan@robots.ox.ac.uk}$
\\
\And
  Sebastian Nowozin\\
Microsoft Research Cambridge\\
$\texttt{sebastian.nowozin@microsoft.com}$
}

\begin{document}

\maketitle
\vspace{-8mm}
\begin{abstract}
We present a new type of probabilistic model which we call DISsimilarity COefficient Networks (DISCO Nets). DISCO Nets allow us to efficiently sample from a posterior distribution parametrised by a neural network. During training, DISCO Nets are learned by minimising the dissimilarity coefficient between the true distribution and the estimated distribution. This allows us to tailor the training to the loss related to the task at hand. We empirically show that (i) by modeling uncertainty on the output value, DISCO Nets outperform equivalent non-probabilistic predictive networks and (ii) DISCO Nets accurately model the uncertainty of the output, outperforming existing probabilistic models based on deep neural networks.
\end{abstract}
\mysection{Introduction}
\label{intro}
We are interested in the class of problems that require the prediction of a structured output~$\bm{y} \in \mathcal{\bm{Y}}$ given an input~$\bm{x} \in \mathcal{\bm{X}}$. Complex applications often have large uncertainty on the correct value of~$\bm{y}$. For example, consider the task of hand pose estimation from depth images, where one wants to accurately estimate the pose~$\bm{y}$ of a hand given a depth image~$\bm{x}$. The depth image often has some occlusions and missing depth values and this results in some uncertainty on the pose of the hand. It is, therefore, natural to use probabilistic models that are capable of representing this uncertainty. Often, the capacity of the model is restricted and cannot represent the true distribution perfectly. In this case, the choice of the learning objective influences final performance. Similar to~\citet{DBLP:journals/jmlr/Lacoste-JulienHG11}, we argue that the learning objective should be tailored to the evaluation loss in order to obtain the best performance with respect to this loss. In details, we denote by~$\Delta_{\text{training}}$ the loss function employed during model training, and by~$\Delta_{\text{task}}$ the loss employed to evaluate the model's performance.\\
\\
We present a simple example to illustrate the point made above. We consider a data distribution that is a mixture of two bidimensional Gaussians. We now consider two models to capture the data probability distribution. Each model is able to represent a bidimensional Gaussian distribution with diagonal covariance parametrised by~$(\mu_1, \mu_2, \sigma_1, \sigma_2)$. In this case, neither of the models will be able to recover the true data distribution since they do not have the ability to represent a mixture of Gaussians. In other words, we cannot avoid model error, similarly to the real data scenario. Each model uses its own training loss~$\Delta_{\text{training}}$. Model A employs a loss that emphasises on the first dimension of the data, specified for~$\bm{x} = (x_1,x_2), \bm{x'} = (x_1',x_2') \in \mathbb{R}^2$ by~$\Delta_{A}(\bm{x} - \bm{x'}) = (10 \times (x_1 - x_1')^2 + 0.1 \times (x_2-x_2')^2)^{\sfrac{1}{2}}$. Model B does the opposite and employs the loss function~$\Delta_{B}(\bm{x} - \bm{x'}) = (0.1 \times (x_1 - x_1')^2 + 10\times (x_2-x_2')^2)^{\sfrac{1}{2}}$. Each model performs a grid search over the best parameters values for~$(\mu_1, \mu_2, \sigma_1, \sigma_2)$. Figure~\ref{contour} shows the contours of the Mixture of Gaussians distribution of the data (in black), and the contour of the Gaussian fitted by each model (in red and green). Detailed setting of this example is available in the supplementary material.\\
\begin{minipage}{.4\textwidth}
\captionof{table}{$\Delta_{\text{task}} \pm$ SEM (standard error of the mean) with respect to $\Delta_{\text{training}}$ employed. Evaluation is done the test set.}
\centering
\resizebox{1\textwidth}{!}{
\setlength{\tabcolsep}{1pt}
\begin{tabular}[t]{c|cc}
 \backslashbox{$\Delta_{\text{training}}$}{$\Delta_{\text{task}}$}& $\Delta_{A}$ & $\Delta_{B}$ \\ \hline
$\Delta_{A}$ & $\bm{11.6 \pm 0.287}$   & $13.7 \pm 0.331$ \\
$\Delta_{B}$  & $12.1 \pm 0.305$  & $\bm{11.0 \pm 0.257}$
\label{toytable}
    \end{tabular}
}
  \end{minipage}
\hfill
 \begin{minipage}{.6\textwidth}
\centering
\includegraphics[width=0.8\textwidth]{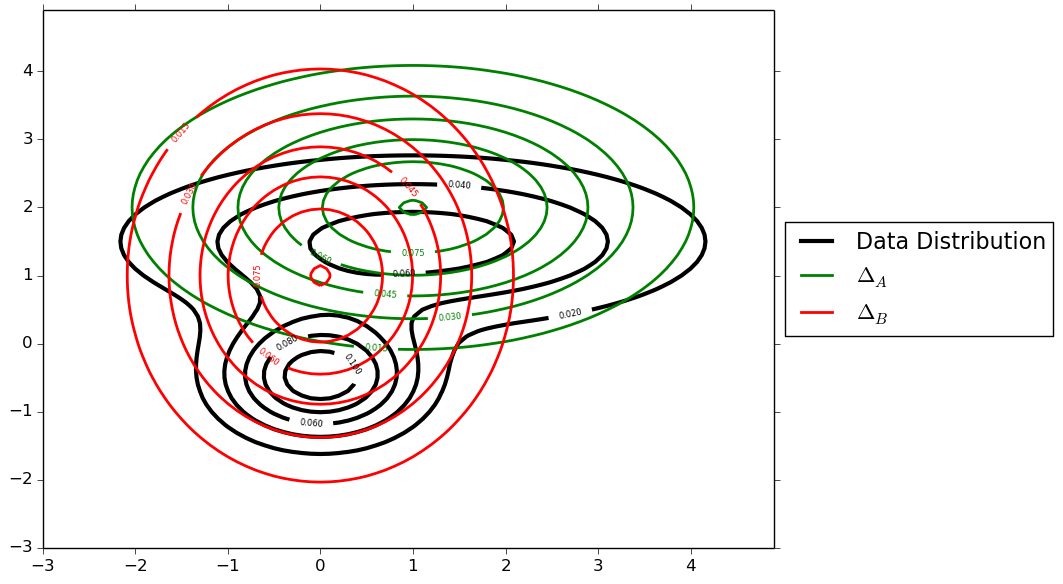}
\captionof{figure}{Contour lines of the Gaussian distribution fitted by each model on the Mixture of Gaussians data distribution. Best viewed in color.}
\label{contour}
\end{minipage}
\\
As expected, the fitted Gaussian distributions differ according to~$\Delta_{\text{training}}$ employed. Table~\ref{toytable} shows that the loss on the test set, evaluated with~$\Delta_{\text{task}}$, is minimised if~$\Delta_{\text{training}} = \Delta_{\text{task}}$. This simple example illustrates the advantage to being able to tailor the model's training objective function to have~$\Delta_{\text{training}} = \Delta_{\text{task}}$. This is in contrast to the commonly employed learning objectives we present in Section~\ref{relatedwork}, that are agnostic of the evaluation loss.\\
\vspace{-2mm}
\\
In order to alleviate the aforementioned deficiency of the state-of-the-art, we introduce DISCO Nets, a new class of probabilistic model. DISCO Nets represent~$P$, the true posterior distribution of the data, with a distribution~$Q$ parametrised by a neural network. We design a learning objective based on a dissimilarity coefficient between~$P$ and~$Q$. The dissimilarity coefficient we employ was first introduced by~\citet{rao} and is defined for any non-negative symmetric loss function. Thus, any such loss can be incorporated in our setting, allowing the user to tailor DISCO Nets to his or her needs. Finally, contrarily to existing probabilistic models presented in Section~$\ref{relatedwork}$, DISCO Nets do not require any specific architecture or training procedure, making them an efficient and easy-to-use class of model.
\vspace{-1mm}
\mysection{Related Work}
\label{relatedwork}
\vspace{-2mm}
Deep neural networks, and in particular, Convolutional Neural Networks (CNNs) are comprised of several convolutional layers, followed by one or more fully connected (dense) layers, interleaved by non-linear function(s) and (optionally) pooling. Recent  probabilistic models use CNNs to represent non-linear functions of the data. We observe that such models separate into two types. The first type of model does not explicitly compute the probability distribution of interest. Rather, these models allow the user to sample from this distribution by feeding the CNN with some noise~$\bm{z}$. Among such models, Generative Adversarial Networks (GAN) presented in~\citet{NIPS2014_5423} are very popular and have been used in several computer vision applications, for example in~\citet{NIPS2015_5773,DBLP:journals/corr/Springenberg15,DBLP:journals/corr/RadfordMC15} and  \citet{DBLP:journals/corr/YanYSL15}. A GAN model consists of two networks, simultaneously trained in an adversarial manner. A generative model, referred as the \emph{Generator} G, is trained to replicate the data from noise, while an adversarial discriminative model, referred as the \emph{Discriminator} D, is trained to identify whether a sample comes from the true data or from G. The GAN training objective is based on a minimax game between the two networks and approximately optimizes a Jensen-Shannon divergence. However, as mentioned in~\citet{NIPS2014_5423} and~\citet{DBLP:journals/corr/RadfordMC15}, GAN models require very careful design of the networks' architecture. Their training procedure is tedious and tends to oscillate. GAN models have been generalized to conditional GAN (cGAN) in~\citet{DBLP:journals/corr/MirzaO14}, where some additional input information can be fed to the \emph{Generator} and the \emph{Discriminator}. For example in~\citet{DBLP:journals/corr/MirzaO14} a cGAN model generates tags corresponding to an image.~\citet{cgans} applies cGAN to face generation.~\citet{RAYLLS16} propose to generate images of flowers with a cGAN model, where the conditional information is a word description of the flower to generate\footnote{At the time writing, we do not have access to the full paper of~\citet{RAYLLS16} and therefore cannot take advantage of this work in our experimental comparison.}. While the application of cGAN is very promising, little quantitative evaluation has been done. Furthermore, cGAN models suffer from the same difficulties we mentioned for GAN. Another line of work has developed towards the use of statistical hypothesis testing to learn probabilistic models. In~\citet{gmmn2} and \citet{gmmn}, the authors propose to train generative deep networks with an objective function based on the Maximum Mean Discrepancy (MMD) criterion. The MMD method (see~\citet{mmd1,mmd2}) is a statistical hypothesis test assessing if two probabilistic distributions are similar. As mentioned in~\citet{gmmn2}, the MMD test can been seen as playing the role of an adversary.\\
\\
The second type of model approximates intractable posterior distributions with use of variational inference. The Variational Auto-Encoders (VAE) presented in~\citet{KingmaW13} is composed of a~\emph{probabilistic encoder} and a~\emph{probabilistic decoder}. The~\emph{probabilistic encoder} is fed with the input~$\bm{x} \in \mathcal{\bm{X}}$ and produces a posterior distribution~$P(\bm{z}|\bm{x})$ over the possible values of noise~$\bm{z}$ that could have generated~$\bm{x}$. The~\emph{probabilistic decoder} learns to map the noise~$\bm{z}$ back to the data space~$\mathcal{\bm{X}}$. The training of VAE uses an objective function based on a Kullback-Leibler Divergence. VAE and GAN models have been combined in \citet{DBLP:journals/corr/MakhzaniSJG15}, where the authors propose to regularise autoencoders with an adversarial network. The adversarial network ensures that the posterior distribution~$P(\bm{z}|\bm{x)}$ matches an arbitrary prior~$P(\bm{z})$.\\
\\
 In hand pose estimation, imagine the user wants to obtain accurate positions of the thumb and the index finger but does not need accurate locations of the other fingers. The task loss~$\Delta_{\text{task}}$ might be based on a weighted L2-norm between the predicted and the ground-truth poses, with high weights on the thumb and the index. Existing probabilistic models cannot be tailored to task-specific losses and we propose the DISsimilarity COefficient Networks (DISCO Nets) to alleviate this deficiency.
\mysection{DISCO Nets}
\label{DISCONETS}
We begin the description of our model by specifying how it can be used to generate samples from the posterior distribution, and how the samples can in turn be employed to provide a pointwise estimate. In the subsequent subsection, we describe how to estimate the parameters of the model.

\begin{figure}[h]
\vspace{-4mm}
\centering
\includegraphics[width=0.75\textwidth,trim={4cm 0.5cm 4cm 1.3cm},clip]{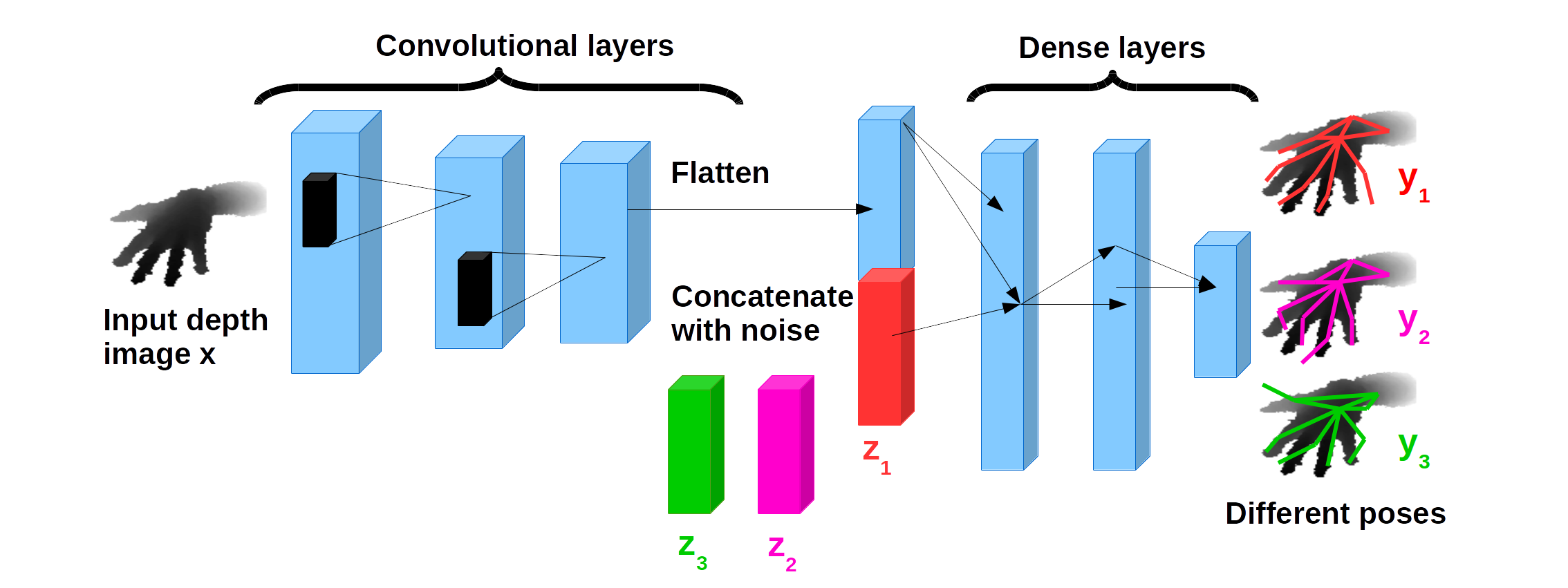}
\captionof{figure}{For a single depth image~$\bm{x}$, using 3 different noise samples~$(\bm{z}_1, \bm{z}_2,\bm{z}_3)$, DISCO Nets output 3 different candidate poses~$(\bm{y}_1,\bm{y}_2,\bm{y}_3)$ (shown superimposed on the depth image). The depth image is from the NYU Hand Pose Dataset of~\citet{tompson14tog}, preprocessed as in~\citet{oberweger15}. Best viewed in color.}
\label{fig::basisarchi}
\vspace{-3mm}
\end{figure}

\mysubsection{Prediction}
\label{model}
\myparagraph{Sampling.}
A DISCO Net consists of several convolutional and dense layers (interleaved by non-linear function(s) and possibly pooling) and takes as input a pair~$(\bm{x},\bm{z}) \in \mathcal{\bm{X}} \times \mathcal{Z}$, where~$\bm{x}$ is input data and~$\bm{z}$ is some random noise. Given one pair~$(\bm{x},\bm{z})$, the DISCO Net produces a value for the output~$\bm{y}$. In the example of hand pose estimation, the input depth image~$\bm{x}$ is fed to the convolutional layers. The output of the last convolutional layer is flattened and concatenated with a noise sample~$\bm{z}$. The resulting vector is fed to several dense layers, and the last dense layer outputs a pose~$\bm{y}$. From a single depth image~$\bm{x}$, by using different noise samples, the DISCO Net produces different pose candidates for the depth image. This process is illustrated in Figure~\ref{fig::basisarchi}. Importantly, DISCO Nets are flexible in the choice of the architecture. For example, the noise could be concatenated at any stage of the network, including at the start.\\
\\
We denote~$Q$ the distribution that is parametrised by the DISCO Net's neural network. For a given input~$\bm{x}$, DISCO Nets provide the user with samples~$\bm{y}$ drawn from~$Q(\bm{y}|\bm{x})$ without requiring the expensive computation of the (often intractable) partition function. In the remainder of the paper we consider~$\bm{x}\in \mathbb{R}^{d_x}, \bm{y}\in \mathbb{R}^{d_y}$ and~$\bm{z}\in \mathbb{R}^{d_z}$.

\myparagraph{Pointwise Prediction.} In order to obtain a single prediction~$\bm{y}$ for a given input $\bm{x}$, DISCO Nets use the principle of Maximum Expected Utility (MEU), similarly to~\citet{export:238251}. The prediction~$\bm{y}_{\Delta_\text{task}}$ maximises the expected utility, or rather minimises the expected task-specific loss~$\Delta_{\text{task}}$, estimated using the sampled candidates. Formally, the prediction is made as follows:
\vspace{-1mm}
\begin{equation} 
\bm{y}_{\Delta_\text{task}} = \argmax_{k \in [1,K]} \text{EU}(\bm{y}_k) = \argmin_{k \in [1,K]} \sum_{k'=1}^{K} \Delta_{\text{task}}(\bm{y}_k, \bm{y}_k')
\end{equation} 
where~$(\bm{y_1}, ... , \bm{y}_K)$ are the candidate outputs sampled for the single input~$\bm{x}$. Details on the MEU method are in the supplementary material.
\mysubsection{Learning DISCO Nets} 
\label{learning}
\vspace{2mm}
\myparagraph{Objective Function.}
We want DISCO Nets to accurately model the true probability~$P(\bm{y}|\bm{x})$ via~$Q(\bm{y}|\bm{x})$. In other words,~$Q(\bm{y}|\bm{x})$ should be as similar as possible to~$P(\bm{y}|\bm{x})$. This similarity is evaluated with respect to the loss specific to the task at hand. Given any non-negative symmetric loss function between two outputs~$\Delta(\bm{y}, \bm{y'})$ with $(\bm{y}, \bm{y'}) \in \mathcal{\bm{Y}} \times \mathcal{\bm{Y}}$, we employ a diversity coefficient that is the expected loss between two samples drawn randomly from the two distributions. Formally, the diversity coefficient is defined as:
\begin{equation}
\textrm{DIV}_{\Delta} (P,Q,D) =   E_{\bm{x} \sim D(\bm{x})} [ E_{\bm{y} \sim P(\bm{y}|\bm{x})} [ E_{\bm{y'} \sim Q(\bm{y'}|\bm{x})} [\Delta(\bm{y}, \bm{y'})]]]
\end{equation}
Intuitively, we should minimise~$\textrm{DIV}_{\Delta} (P,Q,D)$ so that~$Q(\bm{y}|\bm{x})$ is as similar as possible to~$P(\bm{y}|\bm{x})$. However there is uncertainty on the output~$\bm{y}$ to predict for a given~$\bm{x}$. In other words,~$P(\bm{y}|\bm{x})$ is diverse and~$Q(\bm{y}|\bm{x})$ should be diverse as well. Thus we encourage~$Q(\bm{y}|\bm{x})$ to provide sample outputs, for a given~$\bm{x}$,  that are diverse by minimising the following dissimilarity coefficient:
\begin{equation}
\textrm{DISC}_{\Delta} (P,Q,D) = \textrm{DIV}_{\Delta} (P,Q,D) - \gamma \textrm{DIV}_{\Delta} (Q,Q,D)  - (1-\gamma) \textrm{DIV}_{\Delta} (P,P,D)
\end{equation}
with $\gamma \in [0,1]$. The dissimilarity~$\textrm{DISC}_{\Delta} (P,Q,D)$ is the difference between the diversity between~$P$ and $Q$, and an affine combination of the diversity of each distribution, given~$\bm{x} \sim D(\bm{x})$. These coefficients were introduced by~\citet{rao} with~$\gamma=\sfrac{1}{2}$ and used for latent variable models by~\citet{DBLP:conf/icml/KumarPK12}.  We do not need to consider the term~$\textrm{DIV}_{\Delta} (P,P,D)$ as it is a constant in our problem, and thus the DISCO Nets objective function is defined as follows:
\begin{equation} 
F=\textrm{DIV}_{\Delta} (P,Q,D) - \gamma \textrm{DIV}_{\Delta} (Q,Q,D)
\label{eq:trueobj}
\end{equation}
When minimising~$F$, the term~$\gamma \textrm{DIV}_{\Delta} (Q,Q,D)$ encourages~$Q(\bm{y}|\bm{x})$ to be diverse. The value of~$\gamma$ balances between the two goals of~$Q(\bm{y}|\bm{x})$ that are providing accurate outputs while being diverse. 
\myparagraph{Optimisation.}
Let us consider a  training dataset composed of~$N$ examples input-output pairs~$D = \{(\bm{x}_n,\bm{y}_n), n= 1.. N\}$. In order to train DISCO Nets, we need to compute the objective function of equation~\eqref{eq:trueobj}. We do not have knowledge of the true probability distributions~$P(\bm{y},\bm{x})$ and~$P(\bm{x})$. To overcome this deficiency, we construct estimators of each diversity term~$\textrm{DIV}_{\Delta} (P, Q)$ and~$\textrm{DIV}_{\Delta} (Q, Q)$. First, we take an empirical distribution of the data, that is, taking ground-truth pairs~$(\bm{x}_n, \bm{y}_n)$.
We then estimate each distribution~$Q(\bm{y}|\bm{x}_n)$ by sampling~$K$ outputs from our model for each~$\bm{x}_n$. This gives us an unbiased estimate of each diversity term, defined as:
\begin{equation}
\begin{split}
&\widehat{\textrm{DIV}}_{\Delta} (P,Q,D) = \dfrac{1}{N} \sum_{n=1}^{N} \dfrac{1}{K} \sum_{k=1}^{K} \Delta(\bm{y}_n, G(\bm{z}_k,\bm{x}_n;\bm{\theta})) \\
&\widehat{\textrm{DIV}}_{\Delta} (Q,Q,D) =  \dfrac{1}{N} \sum_{n=1}^{N} \dfrac{1}{K(K-1)} \sum_{k=1}^{K} \sum_{k'=1, k' \neq k}^{K} \Delta(G(\bm{z}_k,\bm{x}_n;\bm{\theta}), G(\bm{z}_{k'},\bm{x}_n;\bm{\theta}))
\end{split}
\end{equation}
We have an unbiased estimate of the DISCO Nets' objective function of equation~\eqref{eq:trueobj}:
\begin{equation}
\begin{split}
\widehat{F}(\Delta,\bm{\theta}) = \widehat{\textrm{DIV}}_{\Delta} (P,Q,D) - \gamma \widehat{\textrm{DIV}}_{\Delta} (Q,Q,D)
\label{trainingobj}
\end{split}
\end{equation}
where~$\bm{y}_k = G(\bm{z}_k,\bm{x}_n;\bm{\theta})$ is a candidate output sampled from DISCO Nets for~($\bm{x}_n$,$\bm{z}_k$), and~$\bm{\theta}$ are the parameters of DISCO Nets. It is important to note that the second term of equation~$\eqref{trainingobj}$ is summing over~$k$ and~$k'\neq k$ to have an unbiased estimate, therefore we compute the loss between pairs of different samples~$G(\bm{z}_k,\bm{x}_n; \bm{\theta})$ and $G(\bm{z}_{k'},\bm{x}_n; \bm{\theta})$. The parameters~$\bm{\theta}$ are learned by Gradient Descent. Algorithm~\ref{algo:algtraining} shows the training of DISCO Nets. In steps 4 and 5 of Algorithm~\ref{algo:algtraining}, we draw~$K$ random noise vectors~$(\bm{z}_{n,1}, ... \bm{z}_{n,k})$ per input example~$\bm{x}_n$, and generate~$K$ candidate outputs~$G(\bm{z}_{n,k},\bm{x}_n; \bm{\theta})$ per input. This allow us to compute an unbiased estimate of the gradient in step 7. 
For clarity, in the remainder of the paper we do not explicitely write the parameters~$\bm{\theta}$ and write~$G(\bm{z}_{k},\bm{x}_n)$.
\RestyleAlgo{ruled}
\SetAlgoNoLine
\LinesNumbered
\begin{algorithm}[ht]
\SetAlgoLined
\For{t=1...T epochs}{
Sample minibatch of $b$ training example pairs~$\{(\bm{x}_1, \bm{y}_1)... (\bm{x}_b,\bm{y}_b)\}$.\\
\For{n=1...b}{
Sample $K$ random noise vectors~$(\bm{z}_{n,1}, ... \bm{z}_{n,k})$ for training example~$\bm{x}_n$\\
Generate K candidate outputs~$G(\bm{z}_{n,k},\bm{x}_n; \bm{\theta}), k=1..K$ for training example~$\bm{x}_n$
}
Update parameters~$\bm{\theta}^t \leftarrow \bm{\theta}^{t-1}$ by descending the gradient of equation~\eqref{trainingobj} : $\nabla_{\bm{\theta}} \widehat{F}(\Delta, \bm{\theta})$.
}
\caption{DISCO Nets Training algorithm.}
\label{algo:algtraining}
\end{algorithm}
\mysubsection{Strictly Proper Scoring Rules.}
\label{scorringrule}
\myparagraph{Scoring Rule for Learning.}
A scoring rule~$S(Q,P)$, as defined in~\citet{RePEc:bes:jnlasa:v:102:y:2007:p:359-378}, evaluates the quality of a predictive distribution~$Q$ with respect to a true distribution~$P$. When using a scoring rule one should ensure that it is proper, which means it is maximised when~$P=Q$. A scoring rule is said to be strictly proper if~$P=Q$ is the unique maximiser of~$S$. Hence maximising a proper scoring rule ensures that the model aims at predicting relevant forecast. \citet{RePEc:bes:jnlasa:v:102:y:2007:p:359-378} define score divergences corresponding to a proper scoring rule S:
\vspace{-1mm}
\begin{equation}
d(Q,P) = S(P, P) - S(Q,P)
\vspace{-1mm}
\end{equation}
If~$S$ is proper, ~$d$ is a valid non-negative divergence function, with value 0 if (and only if, in the case of strictly proper)~$Q = P$. For example the MMD criterion (see~\citet{mmd1, mmd2}) mentioned in Section~\ref{relatedwork} is an example of this type of divergence. In our case, any loss~$\Delta$ expressed with an universal kernel will define the DISCO Nets' objective function as such divergence (see~\citet{AAAI159853}). For example, by Theorem 5 of~\citet{RePEc:bes:jnlasa:v:102:y:2007:p:359-378}, if we take as loss function~$\Delta_\beta(\bm{y}, \bm{y'}) = ||\bm{y} - \bm{y'}||_2^\beta = \sum_{i=1}^{d_y} |(y^i - y'^i|^2)^{\sfrac{\beta}{2}}$ with $\beta \in [0,2]$ excluding 0 and 2, our training objective is (the negative of) a strictly proper scoring rule, that is:
\begin{equation}
\resizebox{\textwidth}{!}{$
\widehat{F}(\Delta, \bm{\theta}) = \dfrac{1}{N}\sum_{n=1}^{N} \Big[ \dfrac{1}{K} \sum_{k} ||\bm{y}_n - G(\bm{z}_k,\bm{x}_n)||_{2}^{\beta}- \dfrac{1}{2}\dfrac{1}{K(K-1)} \sum_{k} \sum_{k'\neq k}|| G(\bm{z}_{k'},\bm{x}_n) - G(\bm{z}_k,\bm{x}_n)||_{2}^{\beta}\Big]
$}
\label{objpropscore}
\end{equation}
This score has been referred in the litterature as the Energy Score in~\citet{Gneiting2008, 0ebc934272664672bcf621d3c560d107, RePEc:bes:jnlasa:v:102:y:2007:p:359-378}.\\
\\
By employing a (strictly) proper scoring rule we ensure that our objective function is (only) minimised at the true distribution~$P$, and expect DISCO Nets to generalise better on unseen data. We show below that strictly proper scoring rules are also relevant to assess the quality of the distribution~$Q$ captured by the model. 
\vspace{-2mm}
\myparagraph{Discriminative power of proper scoring rules.}
As observed in~\citet{JMLR:v14:fukumizu13a}, kernel density estimation (KDE) fails in high dimensional output spaces. Our goal is to compare the quality of the distribution captured between two models,~$Q_1$ and $Q_2$. In our setting $Q_1$ better models~$P$ than~$Q_2$ according to the scoring rule~$S$ and its associated divergence~$d$ if $d(Q_1,P) <d(Q_2,P)$. As noted in~\citet{0ebc934272664672bcf621d3c560d107}, $S$ being proper does not ensure~$d(Q_1,\bm{y}) < d(Q_2, \bm{y})$ for all observations~$\bm{y}$ drawn from~$P$. However if the scoring rule is strictly proper scoring rule, this property should be ensured in the neighbourhood of the true distribution. 

\mysection{Experiments : Hand Pose Estimation}
Given a depth image~$\bm{x}$, which often contains occlusions and missing values, we wish to predict the hand pose~$\bm{y}$. We use the NYU Hand Pose dataset of~\citet{tompson14tog} to estimate the efficiency of DISCO Nets for this task. 
\mysubsection{Experimental Setup}
\label{prelim}
\myparagraph{NYU Hand Pose Dataset.}
The NYU Hand Pose dataset of~\citet{tompson14tog} contains 8252 testing and 72,757 training frames of captured RGBD data with ground-truth hand pose information. The training set is composed of images of one person whereas the testing set gathers samples from two persons. For each frame, the RGBD data from 3 Kinects is provided: a frontal view and 2 side views. In our experiments we use only the depth data from the frontal view. While the ground truth contains J = 36 annotated joints, we follow the evaluation protocol of~\citet{oberweger15,oberweger15a} and use the same subset of J = 14 joints. We also perform the same data preprocessing as in~\citet{oberweger15,oberweger15a}, and extract a fixed-size metric cube around the hand from the depth image. We resize the depth values within the cube to a~$128\times128$ patch and normalized them in~$[-1,1]$. Pixels deeper than the back of the cube and missing depth values are both set to a depth of 1. 

\myparagraph{Methods.}
We employ loss functions between two outputs of the form of the Energy score~\eqref{objpropscore}, that is, $\Delta_{\textrm{training}} = \Delta_\beta(\bm{y}, \bm{y'}) = ||\bm{y} - \bm{y'}||_2^\beta$. Our first goal is to assess the advantages of DISCO Nets with respect to non-probabilistic deep networks. One model, referred as~$\text{DISCO}_{\beta, \gamma}$, is a DISCO Nets probabilistic model, with $\gamma \neq 0$ in the dissimilarity coefficient of equation~\eqref{trainingobj}. When taking~$\gamma = 0$, noise is injected and the model capacity is the same as~$\text{DISCO}_{\beta, \gamma\neq0}$. The model~$\text{BASE}_\beta$, is a non-probabilistic model, by taking~$\gamma = 0$ in the objective function of equation~\eqref{trainingobj} and no noise is concatenated. This corresponds to a classic deep network which for a given input~$\bm{x}$ generates a single output~$\bm{y} = G(\bm{x})$. Note that we write~$G(\bm{x})$ and not~$G(\bm{z},\bm{x})$ since no noise is concatenated. 

\myparagraph{Evaluation Metrics.}
We report classic non-probabilistic metrics for hand pose estimation employed in~\citet{oberweger15,oberweger15a} and~\citet{ff}, that are, the Mean Joint Euclidean Error (MeJEE), the Max Joint Euclidean Error (MaJEE) and the Fraction of Frames within distance (FF). We refer the reader to the supplementary material for detailed expression of these metrics. These metrics use the Euclidean distance between the prediction and the ground-truth and require a single pointwise prediction. This pointwise prediction is chosen with the MEU method among~$K$ candidates. We added the probabilistic metric ProbLoss. ProbLoss is defined as in Equation~\ref{objpropscore} with the Euclidean norm and is the divergence associated with a strictly proper scoring rule. Thus, ProbLoss ranks the ability of the models to represent the true distribution. ProbLoss is computed using~$K$ candidate poses for a given depth image. For the non-probabilistic model~$\text{BASE}_{\beta}$, only a single pointwise predicted output~$\bm{y}$ is available. We construct the~$K$ candidates by adding some Gaussian random noise of mean~$0$ and diagonal covariance~$\Sigma = \sigma \mathds{1}$, with~$\sigma \in \{1\textrm{mm}, 5\textrm{mm}, 10\textrm{mm}\}$ and refer to the model as~$\text{BASE}_{\beta, \sigma}$. \footnote{We also evaluate the non-probabilistic model~$\text{BASE}_{\beta}$ using its pointwise prediction rather than the MEU method. Results are consistent and detailed in the supplementary material.} 

\myparagraph{Loss functions.}
As we employ standard evaluation metrics based on the Euclidean norm, we train with the Euclidean norm (that is,~$\Delta_{\textrm{training}}(\bm{y}, \bm{y'}) =  ||\bm{y} - \bm{y'}||_2^\beta$ with $\beta=1$). When~$\gamma=\frac{1}{2}$ our objective function coincides with ProbLoss.

\myparagraph{Architecture.}
The novelty of DISCO Nets resides in their objective function. They do not require the use of a specific network architecture. This allows us to design a simple network architecture inspired by~\citet{oberweger15a}. The architecture is shown in Figure~$\ref{fig::basisarchi}$. The input depth image~$\bm{x}$ is fed to 2 convolutional layers, each having 8 filters, with kernels of size~$5 \times 5$, with stride 1, followed by Rectified Linear Units (ReLUs) and Max Pooling layers of kernel size~$3 \times 3$. A third and last convolutional layer has 8 filters, with kernels of size~$5 \times 5$, with stride 1, followed by a Rectified Linear Unit. The ouput of the convolution is concatenated to the random noise vector~$\bm{z}$ of size~$d_z = 200$, drawn from a uniform distribution in~$[-1,1]$. The result of the concatenation is fed to 2 dense layers of output size 1024, with ReLUs, and a third dense layer that outputs the candidate pose~$\bm{y} \in {\mathbb{R}}^{3 \times J}$. For the non-probabilistic~$\text{BASE}_{\beta, \sigma}$ model no noise is concatenated as only a pointwise estimate is produced.

\begin{figure}[h]
\vspace{-3mm}
\begin{subfigure}[t]{0.3\textwidth}
\centering
\includegraphics[width=0.4\textwidth, trim={12cm 10cm 10cm 8cm},clip]{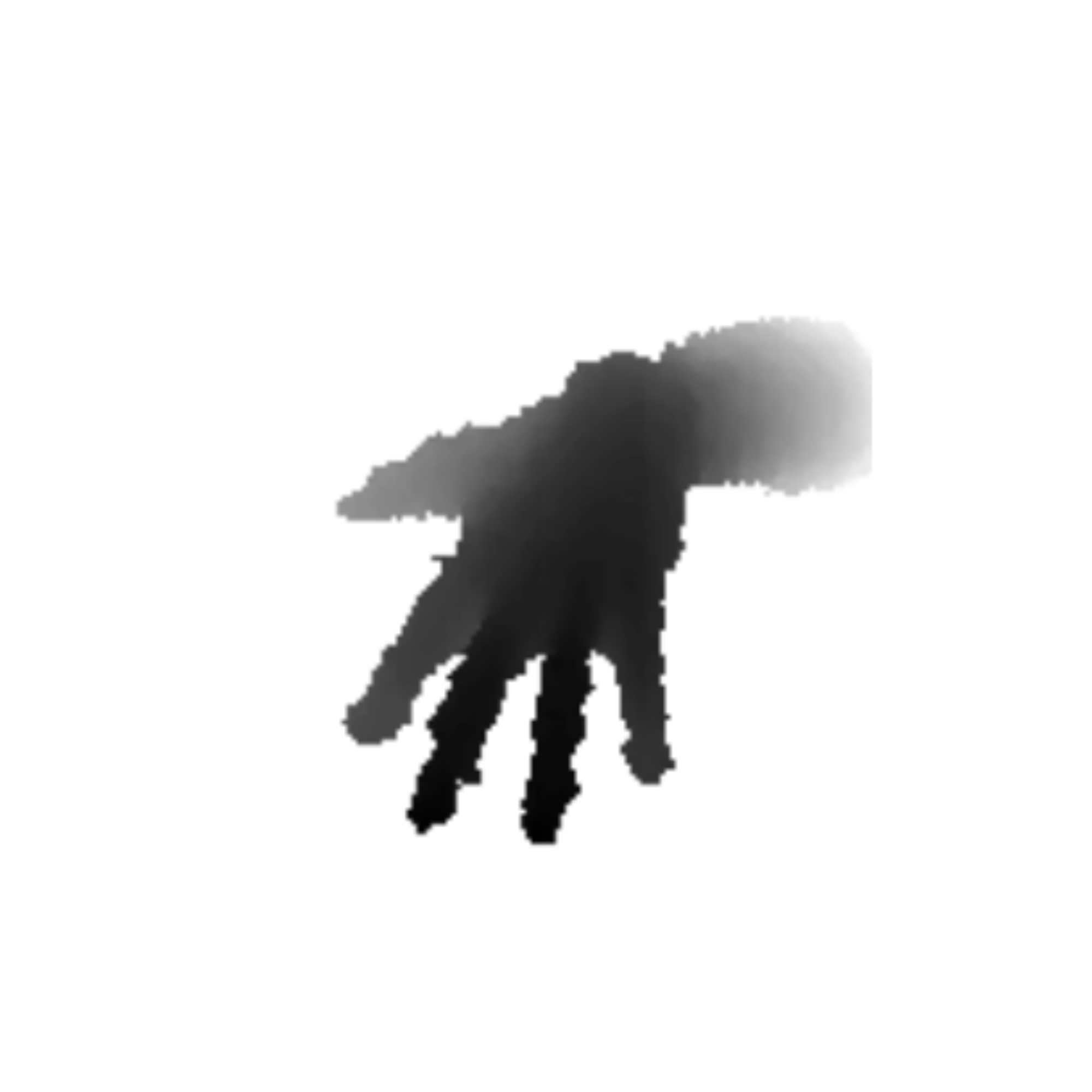}
\includegraphics[width=0.4\textwidth, trim={12cm 10cm 10cm 8cm},clip]{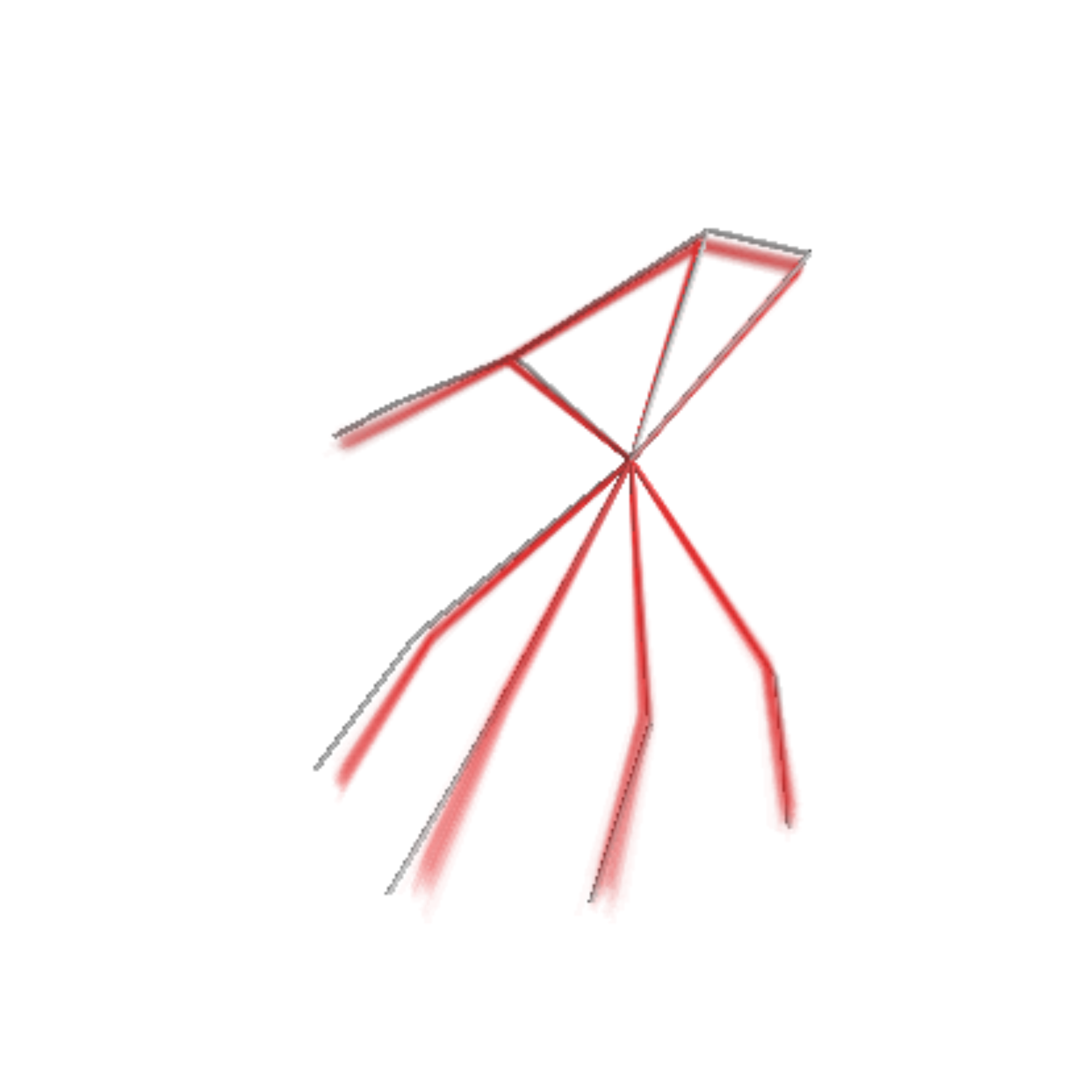}
\mycaption{When there are no occlusions, DISCO Nets model low uncertainty on all joints.}
\end{subfigure}
\hfill
\begin{subfigure}[t]{0.3\textwidth}
\centering
\includegraphics[width=0.4\textwidth, trim={12cm 10cm 10cm 8cm},clip]{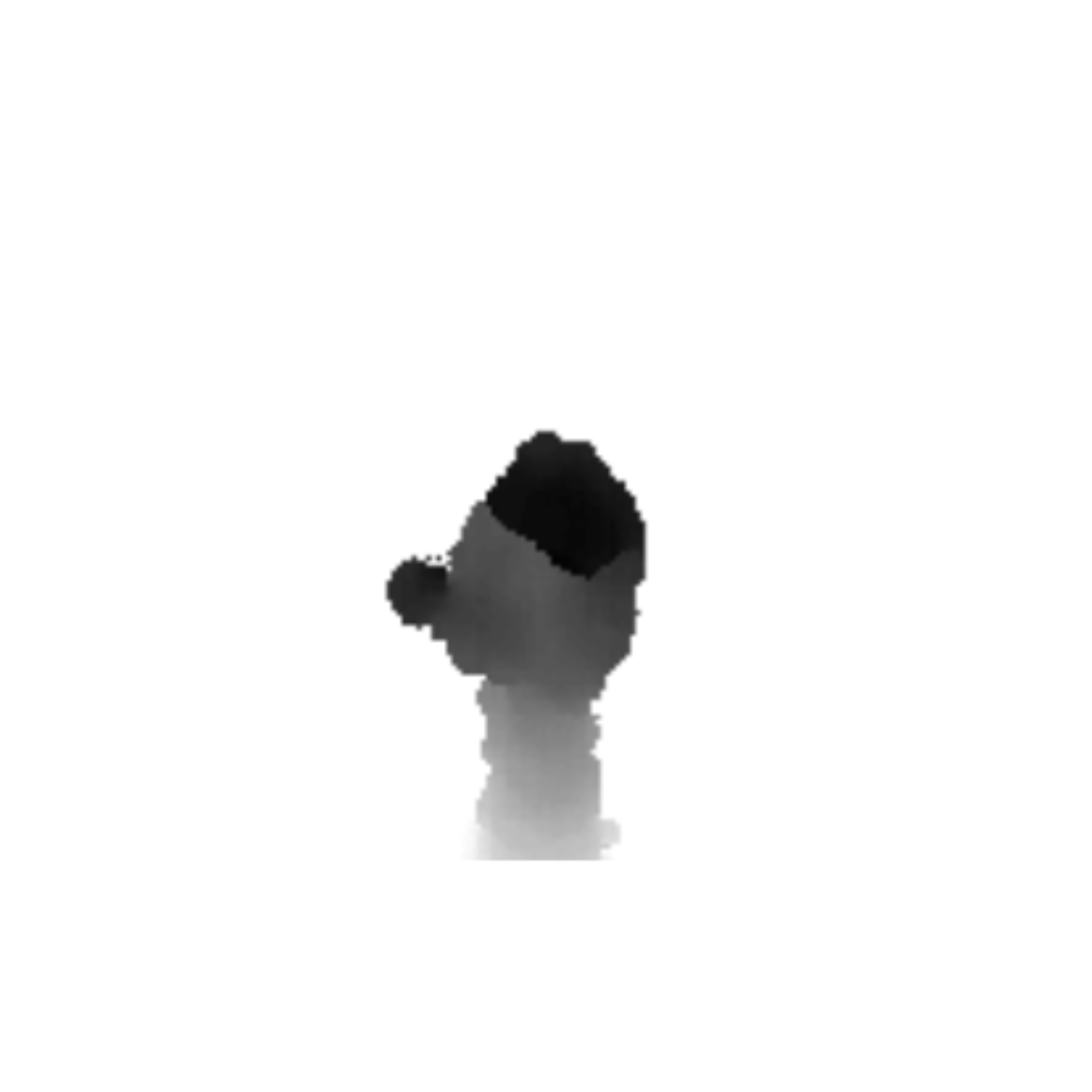}
\includegraphics[width=0.4\textwidth,trim={12cm 10cm 10cm 8cm},clip]{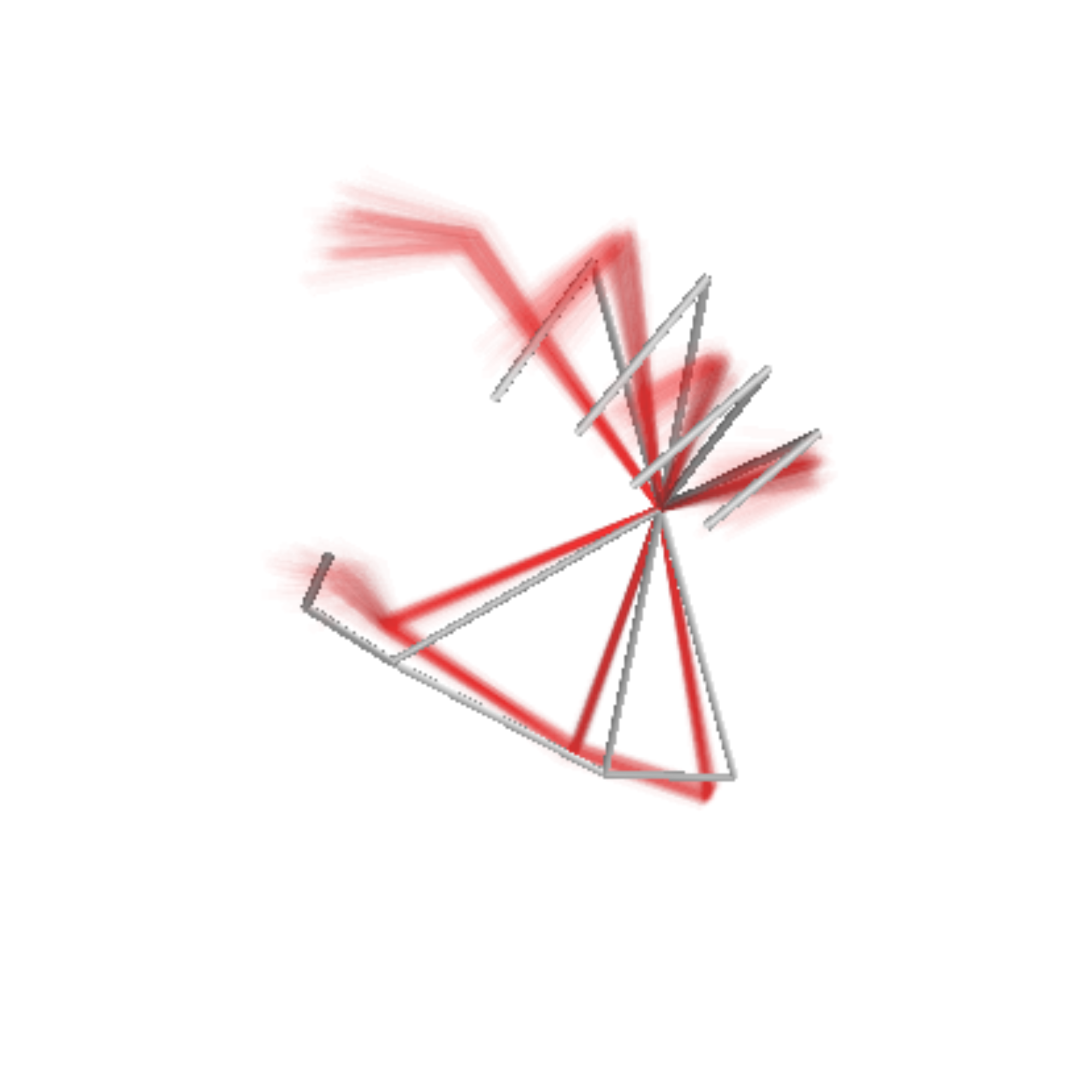}
\mycaption{When the hand is half-fisted, DISCO Nets model the uncertainty on the location of the fingertips.}
\end{subfigure}
\hfill
\begin{subfigure}[t]{0.3\textwidth}
\centering
\includegraphics[width=0.4\textwidth, trim={12cm 10cm 10cm 8cm},clip]{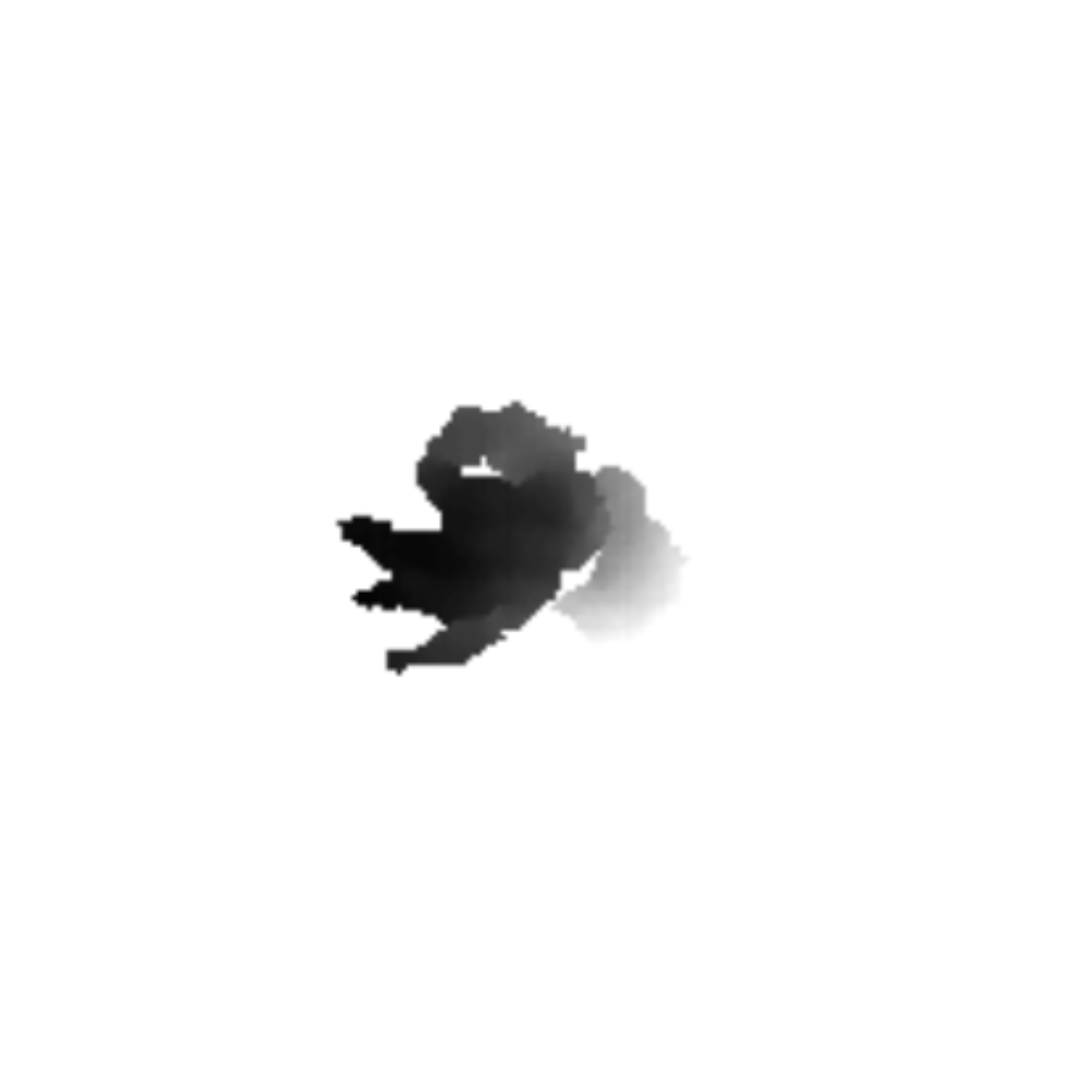}
\includegraphics[width=0.4\textwidth,trim={12cm 10cm 10cm 8cm},clip]{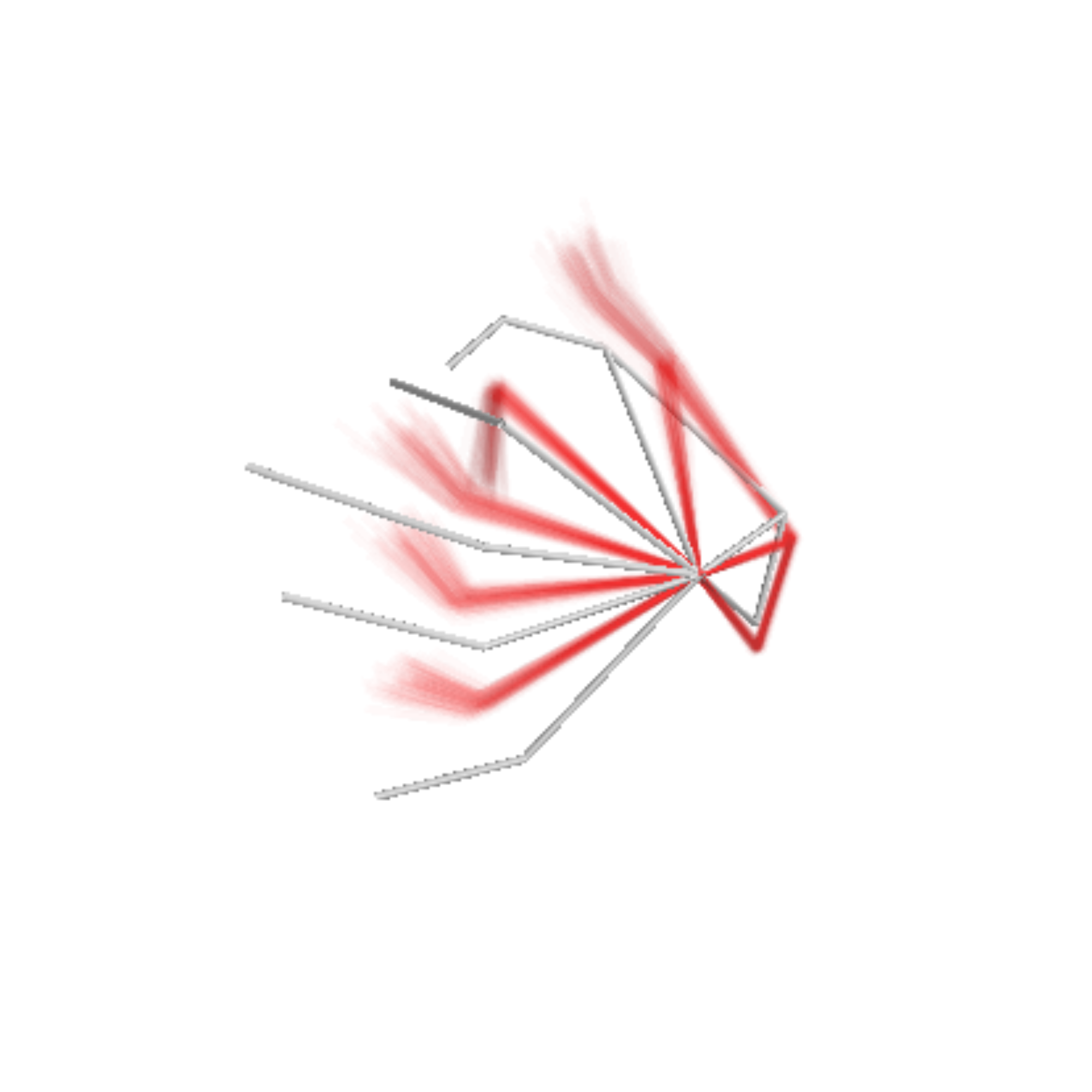}
\mycaption{Here the fingertips of all fingers but the forefinger are occluded and DISCO Nets model high uncertainty on them.}
\end{subfigure}
\vspace{3mm}
\mycaption{Visualisation of~$\text{DISCO}_{\beta = 1, \gamma = 0.5}$ predictions for 3 examples from the testing dataset. The left image shows the input depth image, and the right image shows the ground-truth pose in grey with 100 candidate outputs superimposed in transparent red. Best viewed in color.}
\label{fig::viz}
\end{figure}

\begin{figure}[b]
\begin{minipage}{0.45\textwidth}
\centering
\begin{tikzpicture}
  \node (img)  {\includegraphics[width=0.7\textwidth]{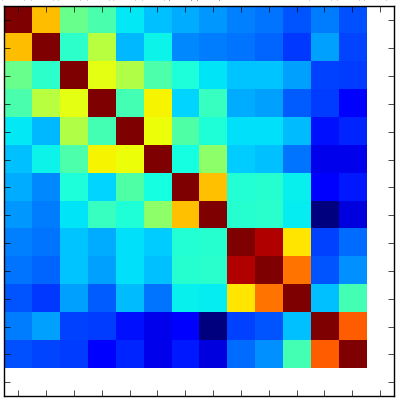}};
 \node (leg) [below=of img, node distance=0cm, yshift=1cm,font=\color{black}, font=\tiny] {P~PR~PL~TR~TM~TT~IM~IT~MM~MT~RM~RT~PM~PT};
 \node[left=of img, node distance=0cm, rotate=90, anchor=center,yshift=-0.7cm,font=\color{black}, font=\tiny] {P~PR~PL~TR~TM~TT~IM~IT~MM~MT~RM~RT~PM~PT};
\node[below=of leg, node distance=0cm, yshift=1cm,font=\color{black}, font=\scriptsize]{$\gamma=0$};
 \end{tikzpicture}
\end{minipage}
\hspace{10mm}
\begin{minipage}{0.45\textwidth}
\centering
\begin{tikzpicture}
  \node (img)  {\includegraphics[width=0.7\textwidth]{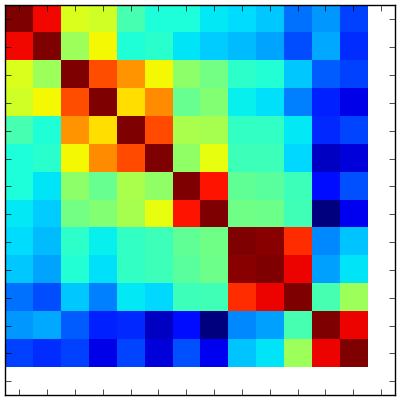}};
 \node (leg) [below=of img, node distance=0cm, yshift=1cm,font=\color{black}, font=\tiny] {P~PR~PL~TR~TM~TT~IM~IT~MM~MT~RM~RT~PM~PT};
 \node[left=of img, node distance=0cm, rotate=90, anchor=center,yshift=-0.7cm,font=\color{black}, font=\tiny] {P~PR~PL~TR~TM~TT~IM~IT~MM~MT~RM~RT~PM~PT};
\node[below=of leg, node distance=0cm, yshift=1cm,font=\color{black}, font=\scriptsize]{$\gamma=0.5$};
\end{tikzpicture}
\end{minipage}
\mycaption{Pearson coefficients matrices of the joints: Palm (no value as the empirical variance is null), Palm Right, Palm Left, Thumb Root, Thumb Mid, Index Mid, Index Tip, Middle Mid, Middle Tip, Ring Mid, Ring Tip, Pinky Mid, Pinky Tip.}
\label{pearsonmatrices}
\end{figure}

\myparagraph{Training.}
We use 10,000 examples from the 72,757 training frames to construct a validation dataset and train only on 62,757 examples. Back-propagation is used with Stochastic Gradient Descent with a batchsize of~$256$. The learning rate is fixed to~$\lambda = 0.01$ and we use a momentum of~$m = 0.9$ (see~\citet{momentum}). We also add L2-regularisation controlled by the parameter~$C$. We use~$C=[0.0001,0.001,0.01]$ which is a relevant range as the comparative model $\text{BASE}_\beta$ is best performing for $C=0.001$. Note that DISCO Nets report consistent performances across the different values~$C$, contrarily to~$\text{BASE}_\beta$. We use 3 different random seeds to initialize each model network parameters. We report the performance of each model with its best cross-validated seed and $C$. We train all models for 400 epochs as it results in a change of less than~$3\%$ in the value of the loss on the validation dataset for~$\text{BASE}_{\beta}$. We refer the reader to the supplementary material for details on the setting. 
\mysubsection{Results.} 
\label{ssec::probadv}
\myparagraph{Quantitative Evaluation.}
Table~\ref{tab1} reports performances on the test dataset, with parameters cross-validated on the validation set. All versions of the DISCO Net model outperform the ~$\text{BASE}_\beta$ model. Among the different values of~$\gamma$, we see that~$\gamma=0.5$ better captures the true distribution (lower ProbLoss) while retaining accurate performance on the standard pointwise metrics. Interestingly, using an all-zero noise at test-time gives similar performances on pointwise metrics. We link this to the observation that both the MEAN and the MEU method perform equivalently on these metrics (see supplementary material).
\vspace{-3mm}
\myparagraph{Qualitative Evaluation.}
In Figure~\ref{fig::viz} we show candidate poses generated by~$\text{DISCO}_{\beta = 1, \gamma = 0.5}$ for 3 testing examples. The left image shows the input depth image, and the right image shows the ground-truth pose (in grey) with 100 candidate outputs (superimposed in transparent red). The model predict the joint locations and we interpolate the joints with edges. If an edge is thinner and more opaque, it means the different predictions overlap and that the uncertainty on the location of the edge's joints is low. We can see that~$\text{DISCO}_{\beta = 1, \gamma = 0.5}$ captures relevant information on the structure of the hand.
\begin{table}[t]
\centering
\setlength{\tabcolsep}{1pt}
\begin{minipage}[t]{0.49\textwidth}
\centering
\caption{Metrics values on the test set~$\pm$ SEM. Best performances in bold.}
\resizebox{\textwidth}{!}{
\begin{tabular}{lcccc}
Model& ProbLoss (mm) & MeJEE (mm) &MaJEE (mm) &FF (80mm) \\\hline
$\text{BASE}_{\beta = 1, \sigma =1}$& 103.8$\pm$0.627 & 25.2$\pm$0.152 & 52.7$\pm$0.290 & 86.040 \\ 
$\text{BASE}_{\beta = 1, \sigma =5}$& 99.3$\pm$0.620  & 25.5$\pm$0.151 & 52.9$\pm$0.289 & 85.773 \\ 
$\text{BASE}_{\beta = 1, \sigma =10}$& 96.3$\pm$0.612 & 25.7$\pm$0.149 & 53.2$\pm$0.288 & 85.664 \\ 
\hline
$\text{DISCO}_{\beta = 1, \gamma=0}$& 92.9$\pm$0.533 & 21.6$\pm$0.128 & 46.0$\pm$0.251 & 92.971 \\
$\text{DISCO}_{\beta = 1, \gamma=0.25}$& 89.9$\pm$0.510 & 21.2$\pm$0.122 & 46.4$\pm$0.252 & 93.262\\
$\text{DISCO}_{\beta = 1, \gamma=0.5}$& \textbf{83.8 $\pm$0.503}  & \textbf{20.9$\pm$0.124} & \textbf{45.1$\pm$0.246} & \textbf{94.438}\\
\label{tab1} 
\end{tabular}
}
\end{minipage}
\vspace{-4mm}
\centering
\setlength{\tabcolsep}{1pt}
\begin{minipage}[t]{0.49\textwidth}
\centering
\caption{Metrics values on the test set~$\pm$ SEM for cGAN.}
\resizebox{\textwidth}{!}{
\begin{tabular}{lcccc}
Model& ProbLoss (mm) & MeJEE (mm) &MaJEE (mm) &FF (80mm) \\\hline
$\text{cGAN}$& 442.7$\pm$0.513 & 109.8$\pm$0.128 & 201.4$\pm$0.320 & 0.000 \\
$\text{cGAN}_{\text{init, fixed}}$& 128.9$\pm$0.480 & 31.8$\pm$0.117 & 64.3$\pm$0.230 & 78.454 \\ 
\label{cgantab} 
\end{tabular}
}
\end{minipage}
\end{table}
\\
Figure~\ref{pearsonmatrices} shows the matrices of Pearson product-moment correlation coefficients between joints. We note that DISCO Net with~$\gamma=0.5$ better captures the correlation between the joints of a finger and between the fingers.
\mysubsection{Comparison with existing probabilistic models.}
\vspace{-2mm}
\label{cgancomp}
To the best of our knowledge the conditional Generative Adversarial Networks (cGAN) from~\citet{DBLP:journals/corr/MirzaO14} has not been applied to pose estimation. In order to compare cGAN to DISCO Nets, several issues must be overcome. First, we must design a network architecture for the Discriminator. This is a first disadvantage of cGAN compared to DISCO Nets which require no adversary. Second, as mentioned in~\citet{NIPS2014_5423} and ~\citet{DBLP:journals/corr/RadfordMC15}, GAN (and thus cGAN) require very careful design of the networks' architecture and training procedure. In order to do a fair comparison, we followed the work in~\citet{DBLP:journals/corr/MirzaO14} and practical advice for GAN presented in \citet{bloggan}. We try (i) $\text{cGAN}$, initialising all layers of D and G randomly, and (ii) $\text{cGAN}_{\text{init, fixed}}$ initialising the convolutional layers of G and D with the trained best-performing~$\text{DISCO}_{\beta = 1, \gamma = 0.5}$ of Section~\ref{ssec::probadv}, and keeping these layers fixed. That is, the convolutional parts of G and D are fixed feature extractors for the depth image. This is a setting similar to the one employed for tag-annotation of images in~\citet{DBLP:journals/corr/MirzaO14}. Details on the setting can be found in the supplementary material. Table~\ref{cgantab} shows that the cGAN model obtains relevant results only when the convolutional layers of G and D are initialised with our trained model and kept fixed, that is~$\text{cGAN}_{\text{init, fixed}}$. These results are still worse than DISCO Nets performances. While there may be a better architecture for cGAN, our experiments demonstrate the difficulty of training cGAN over DISCO Nets.
\vspace{-1mm}
\mysubsection{Reference state-of-the-art values.}
\vspace{-1mm}
We train the best-performing~$\text{DISCO}_{\beta = 1, \gamma = 0.5}$ of Section~\ref{ssec::probadv} on the entire dataset, and compare performances with state-of-the-art methods in~Table~\ref{tabSOTA} and Figure~\ref{sotaff}. These state-of-the-art methods are specifically designed for hand pose estimation. In~\citet{oberweger15} a constrained prior hand model, referred as NYU-Prior, is refined on each hand joint position to increase accuracy, referred as NYU-Prior-Refined. In~\citet{oberweger15a}, the input depth image is fed to a first network NYU-Init, that outputs a pose used to synthesize an image with a second network. The synthesized image is used with the input depth image to derive a pose update. We refer to the whole model as NYU-Feedback. On the contrary, DISCO Nets uses a single network whose architecture is similar to NYU-Prior (without constraining on a pose prior). By accurately modeling the distribution of the pose given the depth image, DISCO Nets obtain comparable performances to NYU-Prior and NYU-Prior-Refined. Without any extra effort, DISCO Nets could be embedded in the presented refinement and feedback methods, possibly boosting state-of-the-art performances.\\
\begin{minipage}{0.35\textwidth}
\captionof{table}{DISCO Nets compared to state-of-the-art performances~$\pm$ SEM.}
\label{tabSOTA}
\centering
\resizebox{\textwidth}{!}{
\begin{tabular}{lcccc}
\hline
Model & MeJEE (mm) &MaJEE (mm) &FF (80mm) \\\hline
NYU-Prior& 20.7$\pm$0.150 & 44.8$\pm$0.289 & 91.190 \\ 
NYU-Prior-Refined&  19.7$\pm$0.157 & 44.7$\pm$0.327 & 88.148 \\
NYU-Init& 27.4$\pm$0.152 & 55.4$\pm$0.265 & 86.537 \\ 
NYU-Feedback&16.0$\pm$0.096 & 36.1$\pm$0.208 & 97.334 \\ \hline
$\text{DISCO}_{\beta = 1, \gamma = 0.5}$& 20.7$\pm$0.121 & 45.1$\pm$0.246 & 93.250
\end{tabular}
}
  \end{minipage}
 \begin{minipage}{.65\textwidth}
\centering
\includegraphics[width=0.7\textwidth]{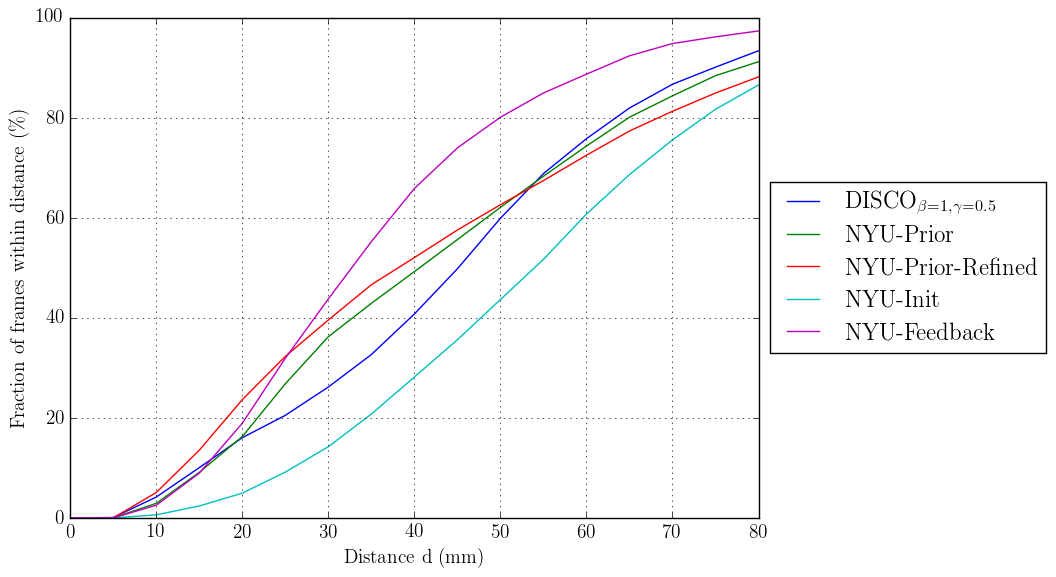}
\vspace{-3mm}
\captionof{figure}{Fractions of frames within distance~$d~\textrm{in mm}$ (by 5 mm). Best viewed in color.}
\label{sotaff}
\end{minipage}
\vspace{-5mm}
\mysection{Discussion.}
We presented DISCO Nets, a new family of probabilistic model based on deep networks. DISCO Nets employ a prediction and training procedure based on the minimisation of a dissimilarity coefficient. Theoretically, this ensures that DISCO Nets accurately capture uncertainty on the correct output to predict given an input. Experimental results on the task of hand pose estimation consistently support our theoretical hypothesis as DISCO Nets outperform non-probabilistic equivalent models, and existing probabilistic models. Furthermore, DISCO Nets can be tailored to the task to perform. This allows a possible user to train them to tackle different problems of interest. As their novelty resides mainly in their objective function, DISCO Nets do not require any specific architecture and can be easily applied to new problems. We contemplate several directions for future work. First, we will apply DISCO Nets to other prediction problems where there is uncertainty on the output. Second, we would like to extend DISCO Nets to latent variables models, allowing us to apply DISCO Nets to diverse dataset where ground-truth annotations are missing or incomplete. 
\mysection{Acknowlegements.}
This work is funded by the Microsoft Research PhD Scholarship Programme. We would like to thank Pankaj Pansari, Leonard Berrada and Ondra Miksik for their useful discussions and insights.
\newpage
\setlength{\bibsep}{0.2pt}
\bibliography{extended_refs.bib}

\begin{thebibliography}{29}
\providecommand{\natexlab}[1]{#1}
\providecommand{\url}[1]{\texttt{#1}}
\expandafter\ifx\csname urlstyle\endcsname\relax
  \providecommand{\doi}[1]{doi: #1}\else
  \providecommand{\doi}{doi: \begingroup \urlstyle{rm}\Url}\fi

\bibitem[Denton et~al.(2015)Denton, Chintala, Szlam, and Fergus]{NIPS2015_5773}
E.L. Denton, S.~Chintala, A.~Szlam, and R.~Fergus.
\newblock Deep generative image models using a {L}aplacian pyramid of
  adversarial networks.
\newblock In \emph{NIPS}. 2015.

\bibitem[Dziugaite et~al.(2015)Dziugaite, Roy, and Ghahramani]{gmmn2}
G.~K. Dziugaite, D.~M. Roy, and Z.~Ghahramani.
\newblock Training generative neural networks via maximum mean discrepancy
  optimization.
\newblock In \emph{UAI}, 2015.

\bibitem[Fukumizu et~al.(2013)Fukumizu, Song, and
  Gretton]{JMLR:v14:fukumizu13a}
K.~Fukumizu, L.~Song, and A.~Gretton.
\newblock Kernel {B}ayes' rule: Bayesian inference with positive definite
  kernels.
\newblock \emph{JMLR}, 2013.

\bibitem[Gauthier(2014)]{cgans}
J.~Gauthier.
\newblock Conditional generative adversarial nets for convolutional face
  generation.
\newblock \emph{Class Project for Stanford CS231N: Convolutional Neural
  Networks for Visual Recognition}, 2014.

\bibitem[Gneiting and Raftery(2007)]{RePEc:bes:jnlasa:v:102:y:2007:p:359-378}
T.~Gneiting and A.~E. Raftery.
\newblock Strictly proper scoring rules, prediction, and estimation.
\newblock \emph{Journal of the American Statistical Association}, 2007.

\bibitem[Gneiting et~al.(2008)Gneiting, Stanberry, Grimit, Held, and
  Johnson]{Gneiting2008}
Tilmann Gneiting, Larissa~I. Stanberry, Eric~P. Grimit, Leonhard Held, and
  Nicholas~A. Johnson.
\newblock Assessing probabilistic forecasts of multivariate quantities, with an
  application to ensemble predictions of surface winds.
\newblock \emph{TEST}, 2008.

\bibitem[Goodfellow et~al.(2014)Goodfellow, Pouget-Abadie, Mirza, Xu,
  Warde-Farley, Ozair, Courville, and Bengio]{NIPS2014_5423}
I.~J. Goodfellow, J.~Pouget-Abadie, M.~Mirza, Bing Xu, D.~Warde-Farley,
  S.~Ozair, A.~Courville, and Y.~Bengio.
\newblock Generative adversarial nets.
\newblock In \emph{NIPS}. 2014.

\bibitem[Gretton et~al.(2007)Gretton, Borgwardt, Rasch, Scholkopf, and
  Smola]{mmd1}
A.~Gretton, K.~M. Borgwardt, M.~J. Rasch, B.~Scholkopf, and A.~J. Smola.
\newblock A kernel method for the two-sample problem.
\newblock In \emph{NIPS}, 2007.

\bibitem[Gretton et~al.(2012)Gretton, Borgwardt, Rasch, Scholkopf, and
  Smola]{mmd2}
A.~Gretton, K.~M. Borgwardt, M.~J. Rasch, B.~Scholkopf, and A.~J. Smola.
\newblock A kernel two-sample test.
\newblock In \emph{JMLR}, 2012.

\bibitem[Kingma and Welling(2014)]{KingmaW13}
D.~P. Kingma and M.~Welling.
\newblock Auto-encoding variational {B}ayes.
\newblock In \emph{ICLR}, 2014.

\bibitem[Kumar et~al.(2012)Kumar, Packer, and Koller]{DBLP:conf/icml/KumarPK12}
M.~P. Kumar, B.~Packer, and D.~Koller.
\newblock Modeling latent variable uncertainty for loss-based learning.
\newblock In \emph{ICML}, 2012.

\bibitem[Lacoste{-}Julien et~al.(2011)Lacoste{-}Julien, Huszar, and
  Ghahramani]{DBLP:journals/jmlr/Lacoste-JulienHG11}
S.~Lacoste{-}Julien, F.~Huszar, and Z.~Ghahramani.
\newblock Approximate inference for the loss-calibrated {B}ayesian.
\newblock In \emph{AISTATS}, 2011.

\bibitem[Larsen and Sønderby()]{bloggan}
A.~B.~L. Larsen and S.~K. Sønderby.
\newblock URL \url{http://torch.ch/blog/2015/11/13/gan.html}.

\bibitem[Li et~al.(2015)Li, Swersky, and Zemel]{gmmn}
Y.~Li, K.~Swersky, and R.~Zemel.
\newblock Generative moment matching networks.
\newblock In \emph{ICML}, 2015.

\bibitem[Makhzani et~al.(2015)Makhzani, Shlens, Jaitly, and
  Goodfellow]{DBLP:journals/corr/MakhzaniSJG15}
A.~Makhzani, J.~Shlens, N.~Jaitly, and I.~J. Goodfellow.
\newblock Adversarial autoencoders.
\newblock \emph{ICLR Workshop}, 2015.

\bibitem[Mirza and Osindero(2014)]{DBLP:journals/corr/MirzaO14}
M.~Mirza and S.~Osindero.
\newblock Conditional generative adversarial nets.
\newblock In \emph{NIPS Deep Learning Workshop}, 2014.

\bibitem[Oberweger et~al.(2015{\natexlab{a}})Oberweger, Wohlhart, and
  Lepetit]{oberweger15}
M.~Oberweger, P.~Wohlhart, and V.~Lepetit.
\newblock Hands deep in deep learning for hand pose estimation.
\newblock In \emph{{Computer Vision Winter Workshop}}, 2015{\natexlab{a}}.

\bibitem[Oberweger et~al.(2015{\natexlab{b}})Oberweger, Wohlhart, and
  Lepetit]{oberweger15a}
M.~Oberweger, P.~Wohlhart, and V.~Lepetit.
\newblock {Training a Feedback Loop for Hand Pose Estimation}.
\newblock In \emph{{ICCV}}, 2015{\natexlab{b}}.

\bibitem[Pinson and Tastu(2013)]{0ebc934272664672bcf621d3c560d107}
Pierre Pinson and Julija Tastu.
\newblock Discrimination ability of the energy score.
\newblock \emph{Technical University of Denmark. (DTU Compute-Technical
  Report-2013; No. 15)}, 2013.

\bibitem[Polyak(1964)]{momentum}
B.~T. Polyak.
\newblock Some methods of speeding up the convergence of iteration methods.
\newblock 1964.

\bibitem[Premachandran et~al.(2014)Premachandran, Tarlow, and
  Batra]{export:238251}
V.~Premachandran, D.~Tarlow, and D.~Batra.
\newblock Empirical minimum {B}ayes risk prediction: How to extract an extra
  few{\%} performance from vision models with just three more parameters.
\newblock In \emph{CVPR}, 2014.

\bibitem[Radford et~al.(2015)Radford, Metz, and
  Chintala]{DBLP:journals/corr/RadfordMC15}
A.~Radford, L.~Metz, and S.~Chintala.
\newblock Unsupervised representation learning with deep convolutional
  generative adversarial networks.
\newblock In \emph{ICLR}, 2015.

\bibitem[Rao(1982)]{rao}
C.R. Rao.
\newblock Diversity and dissimilarity coefficients: A unified approach.
\newblock \emph{Theoretical Population Biology}, pages Vol. 21, No. 1, pp
  24–43, 1982.

\bibitem[Reed et~al.(2016)Reed, Akata, Yan, Logeswaran, Lee, and
  Schiele]{RAYLLS16}
S.~Reed, Z.~Akata, X.~Yan, L.~Logeswaran, H.~Lee, and B.~Schiele.
\newblock Generative adversarial text to image synthesis.
\newblock In \emph{ICML}, 2016.

\bibitem[Springenberg(2016)]{DBLP:journals/corr/Springenberg15}
J.~T. Springenberg.
\newblock Unsupervised and semi-supervised learning with categorical generative
  adversarial networks.
\newblock \emph{ICLR}, 2016.

\bibitem[Taylor et~al.(2012)Taylor, Shotton, Sharp, and Fitzgibbon]{ff}
J.~Taylor, J.~Shotton, T.~Sharp, and A.~Fitzgibbon.
\newblock The vitruvian {M}anifold: Inferring dense correspondences for oneshot
  human pose estimation.
\newblock In \emph{CVPR}, 2012.

\bibitem[Tompson et~al.(2014)Tompson, Stein, Lecun, and Perlin]{tompson14tog}
J.~Tompson, M.~Stein, Y.~Lecun, and K.~Perlin.
\newblock Real-time continuous pose recovery of human hands using convolutional
  networks.
\newblock \emph{ACM Transactions on Graphics}, 2014.

\bibitem[Yan et~al.(2016)Yan, Yang, Sohn, and Lee]{DBLP:journals/corr/YanYSL15}
X.~Yan, J.~Yang, K.~Sohn, and H.~Lee.
\newblock Attribute2image: Conditional image generation from visual attributes.
\newblock 2016.

\bibitem[Zawadzki and Lahaie(2015)]{AAAI159853}
E.~Zawadzki and S.~Lahaie.
\newblock Nonparametric scoring rules.
\newblock In \emph{AAAI Conference on Artificial Intelligence}. 2015.

\end{thebibliography}
\bibliographystyle{plainnat}
\end{document}